\def\year{2022}\relax
\documentclass[letterpaper]{article} 
\usepackage{aaai22}  
\usepackage{times}  
\usepackage{helvet}  
\usepackage{courier}  
\usepackage[hyphens]{url}  
\usepackage{graphicx} 
\usepackage{natbib}  
\usepackage{caption} 
\DeclareCaptionStyle{ruled}{labelfont=normalfont,labelsep=colon,strut=off} 
\def\eg{\emph{e.g.}}
\def\ie{\emph{i.e.}}
\def\etal{\emph{et~al.}}
\usepackage{algorithm}
\usepackage{algorithmic}
\usepackage{amsmath}
\usepackage{bbding}
\usepackage{multirow}
\usepackage{booktabs}
\usepackage{newfloat}
\usepackage{amssymb}
\usepackage{makecell}
\usepackage{listings}
\floatstyle{ruled}
\newfloat{listing}{tb}{lst}{}
\floatname{listing}{Listing}

\title{Content-Variant Reference Image Quality Assessment via \\ Knowledge Distillation}
\author{
	Guanghao Yin\textsuperscript{\rm 1}\thanks{This work was performed while Guanghao Yin worked as an intern at ByteDance.},
	Wei Wang\textsuperscript{\rm 2}\thanks{Equal Contribution.},
	Zehuan Yuan\textsuperscript{\rm 2},
	Chuchu Han\textsuperscript{\rm 3},\\
	Wei Ji\textsuperscript{\rm 4},
	Shouqian Sun\textsuperscript{\rm 1},
	Changhu Wang\textsuperscript{\rm 2},
	\\
}
\affiliations{
	\textsuperscript{\rm 1}
	College of Computer Science and Technology, Zhejiang University, Hangzhou, China,\\
	\textsuperscript{\rm 2} Bytedance Inc, China,
	\textsuperscript{\rm 3} Huazhong University of Science and Technology, Wuhan, China, \\
	\textsuperscript{\rm 4}National University of Singapore, Singapore\\
	{\{ygh\_zju, ssq\}@zju.edu.cn}, {\{wangwei.frank, yuanzehuan,changhu.wang\}@bytedance.com}, hcc@hust.edu.cn, jiwei@nus.edu.sg
}

\usepackage{bibentry}

\begin{document}

\maketitle

\begin{abstract}

Generally, humans are more skilled at perceiving differences between high-quality (HQ) and low-quality (LQ) images than directly judging the quality of a single LQ image. This situation also applies to image quality assessment (IQA). Although recent no-reference (NR-IQA) methods have made great progress to predict image quality free from the reference image, they still have the potential to achieve better performance since HQ image information is not fully exploited. In contrast, full-reference (FR-IQA) methods tend to provide more reliable quality evaluation, but its practicability is affected by the requirement for pixel-level aligned reference images. To address this, we firstly propose the content-variant reference method via knowledge distillation (CVRKD-IQA). Specifically, we use non-aligned reference (NAR) images to introduce various prior distributions of high-quality images. The comparisons of distribution differences between HQ and LQ images can help our model better assess the image quality. Further, the knowledge distillation transfers more HQ-LQ distribution difference information from the FR-teacher to the NAR-student and stabilizing CVRKD-IQA performance. Moreover, to fully mine the local-global combined information,  while achieving faster inference speed, our model directly processes multiple image patches from the input with the MLP-mixer. Cross-dataset experiments verify that our model can outperform all NAR/NR-IQA SOTAs, even reach comparable performance with FR-IQA methods on some occasions. Since the content-variant and non-aligned reference HQ images are easy to obtain, our model can support more IQA applications with its relative robustness to content variations. Our code and more detail elaborations of supplement are available: \url{https://github.com/guanghaoyin/CVRKD-IQA}.
\end{abstract}

\section{Introduction}
The target of objective image quality assessment (IQA) is to quantify the visual distortion and produce the perceptive quality score of the image. The accurate IQA method is quite important to guide many downstream tasks of image processing, such as image restoration~\cite{banham1997digital}, super-resolution~\cite{dong2015image}, etc.

\begin{figure}[tbp]
\centering
\includegraphics[width=8cm]{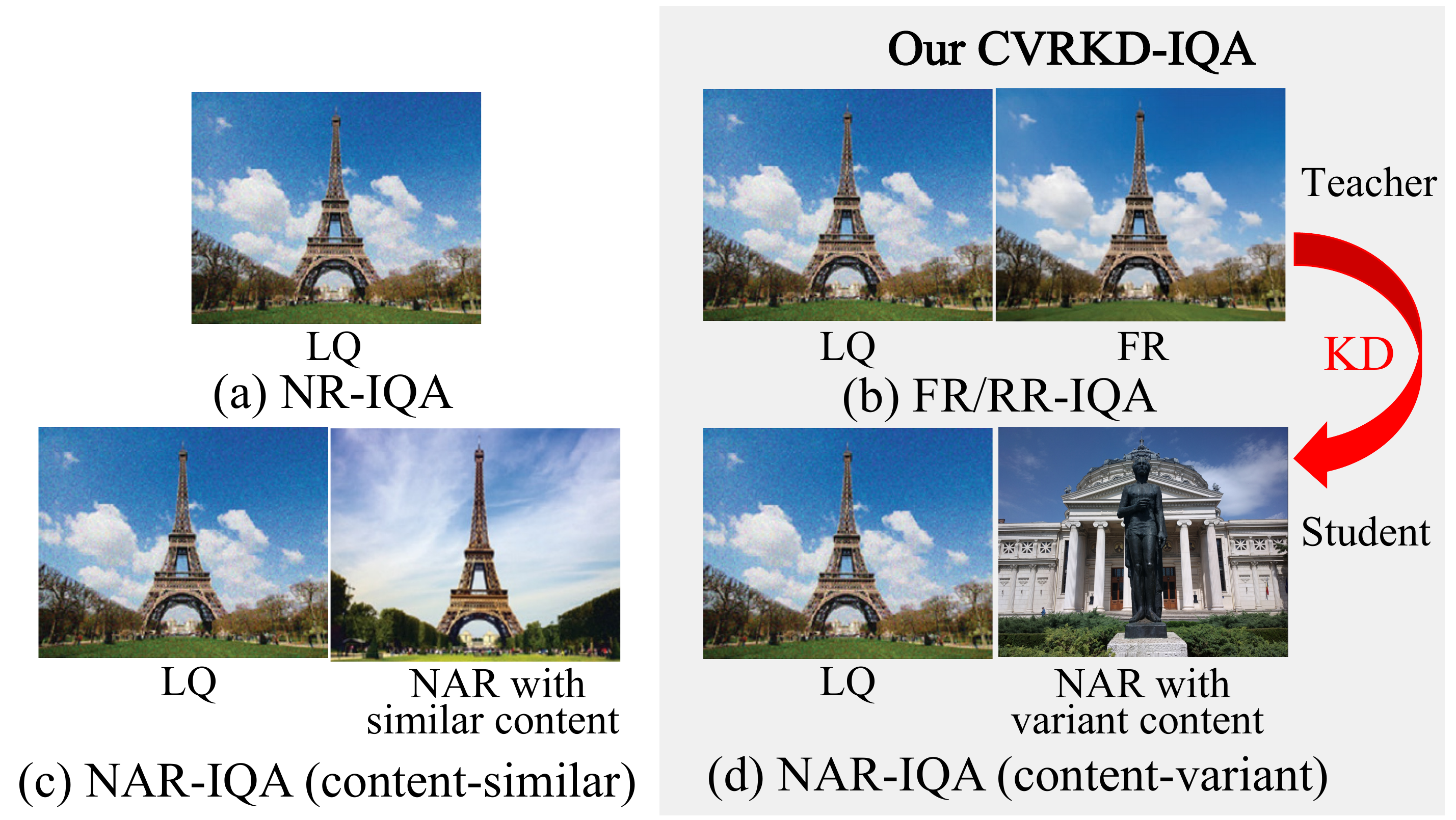}
\caption{Based on the way of using HQ reference images, previous IQA methods are divided into (a) NR-IQA, (b) FR/RR-IQA, (c) NAR-IQA (content-similar). We propose the CVRKD-IQA method, which firstly uses the knowledge distillation to transfer HQ-LQ distribution difference information from (b) FR-IQA to (d) NAR-IQA (content-variant).}
\label{motivation}
\end{figure}

Recent studies on the human visual system~\cite{sheikh2006image, ponomarenko2009tid2008} have shown that humans tend to compare images than directly judging an image. Provided with a high-quality (HQ) reference image, humans can make a more accurate and consistent evaluation about the quality of the distorted image~\cite{ponomarenko2009tid2008}. Based on the way of using HQ reference images, IQA methods are generally divided into three types: no-reference (NR) IQA, reduced-reference (RR) IQA, and full-reference (FR) IQA. Specifically, NR-IQA methods~\cite{bosse2017deep,su2020blindly} (Fig.~\ref{motivation}(a)) only use LQ images as input to directly measure image quality. FR/RR-IQA methods~\cite{rehman2012reduced,cheon2021perceptual} (Fig.~\ref{motivation}(b)) utilize the complete or partial information of the pixel-aligned HQ reference images. Moreover, previous DCNN~\cite{liang2016image} uses the non-aligned image for IQA reference (NAR-IQA), which have similar contents but are not pixel-aligned with the LQ image (Fig.~\ref{motivation}(c)).

Recently, there are several different attempts for NR-IQA methods to achieve promising performance, such as involving a larger-scale database~\cite{lin2019kadid} or using a pretrained feature extractor~\cite{su2020blindly}. Those NR-IQA methods still have the potential for better performance, because they focus more on mining the quality features of LQ image to better fit the labeled scores, but access little HQ image information. Humans can directly judge the quality of one image because they have learned various prior knowledge of HQ-LQ distribution differences before. Hence, we consider involving more explicit HQ prior distribution in the training and inference phases. As proved by previous works~\cite{ponomarenko2009tid2008, liang2016image}, the IQA scores predicted by reference-based IQA methods tend to be more consist with humans than those of NR-IQA methods. However, one strong requirement limits the application of previous FR/NAR-IQA models: their reference images must be pixel-wise aligned or have similar contents with LQ image, which are often unavailable in real scenarios. Thus, on the one hand, we attempt to loose this strong restriction and use content-variant HQ images for reference, since HQ images are available anywhere. On the other hand, inspired by the recent success of cross-modal knowledge distillation (KD)~\cite{lan2018knowledge,porrello2020robust}, we consider to transfer more HQ-LQ difference information from FR-IQA model to NAR-IQA model via KD, which helps NAR-IQA model achieve more accurate and stable performance.

\begin{figure*}[htbp]
\centering
\includegraphics[width=15cm]{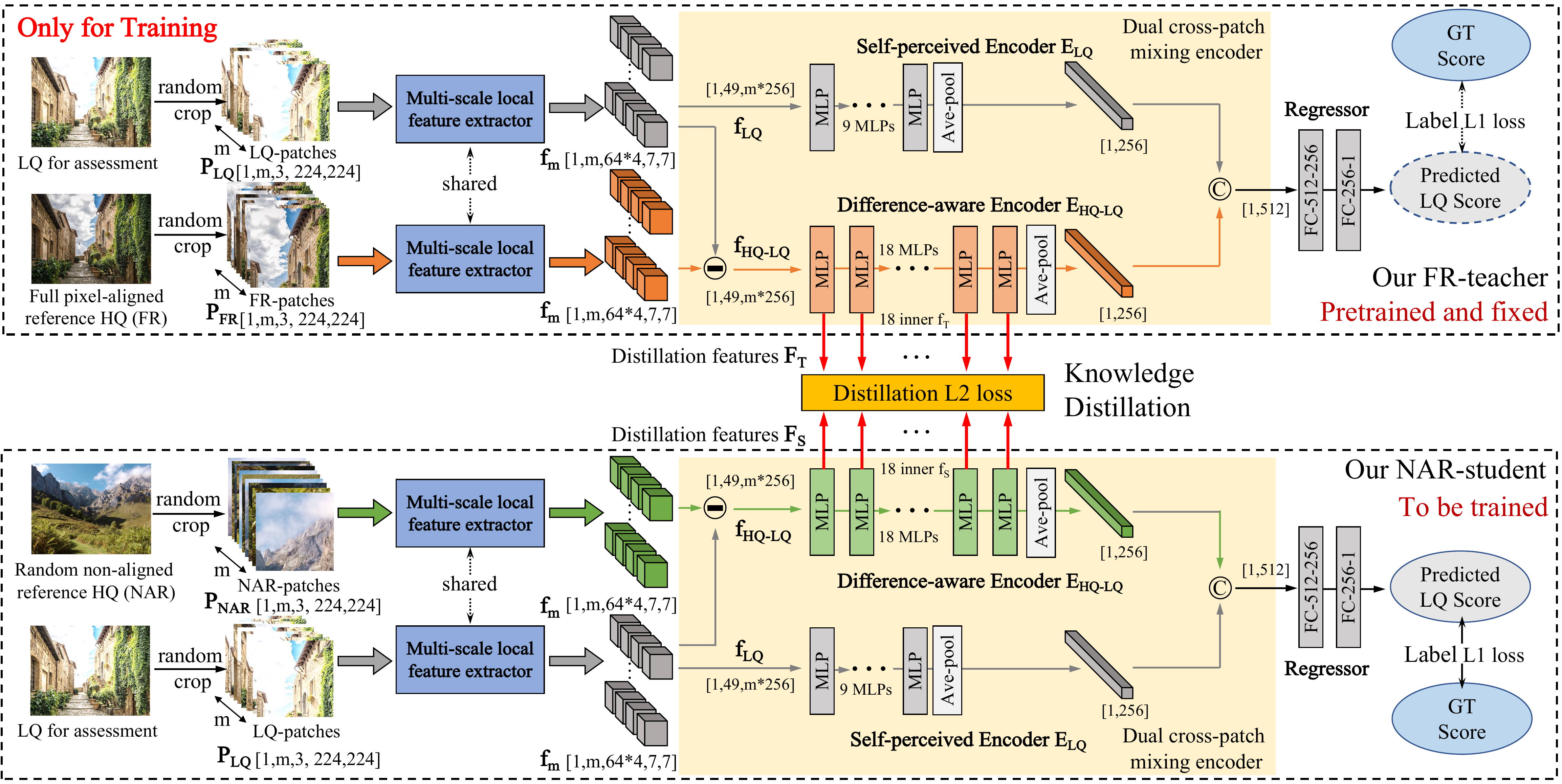}
\caption{Model overview of our CVRKD-IQA. It consists of FR-teacher and NAR-student with the same structure. For better conducting local-global quality description, we use the multi-patches randomly cropped from LQ and reference images as inputs. Note that the FR-teacher is pretrained and fixed only for distillation and the trained NAR-student is applied for testing.}
\label{network}
\end{figure*}

In this paper, we propose the first content-variant reference method via knowledge distillation (CVRKD-IQA) to assess image quality. The structure of our CVRKD-IQA is shown in Fig.~\ref{network}. It consists of two parts: the FR-teacher and NAR-student. They have the same network structure while using different HQ reference images, \ie, the pixel-aligned FR images and content-variant NAR images. For each network branch, the dual-path encoder separately extracts discriminative vectors from the LQ image itself and the HQ-LQ distribution difference. To transfer distribution difference knowledge from FR-teacher to NAR-student, we apply the offline knowledge distillation. The knowledge distillation can also constrain the NAR-student to focus more on useful HQ-LQ distribution difference representation by learning from FR-teacher, and reduce the impact of reference image changes to stabilize the NAR-IQA performance. Specifically, we first train the FR-teacher and fix its parameters, then the layers of the FR-teacher are employed to guide the training of NAR-student. Moreover, to effectively mine the global and local information of the image, our model directly processes a fixed number of image patches sampled from the full image with the classic MLP-mixer~\cite{tolstikhin2021mlp}, which also keeps faster network inference. It should be noted that the FR-teacher is only for training and the NAR-student is applied for testing.

We have conducted extensive comparisons between our model and FR/NR/NAR-IQA SOTAs. Experimental results show that not only our FR-teacher can produce  accurate IQA scores, but also our NAR-student can significantly outperform existing NR/NAR-IQA methods, especially on the large-scale real IQA dataset. On some occasions, our NAR-student can reach comparable performance with some common FR-IQA methods, such as PSNR and LPIPS. It further demonstrates that the proposed strategy transfers HQ-LQ difference prior knowledge from FR-teacher to NAR-student. Moreover, when using different content-variant HQ images, our NAR-student can still keep the relatively stable performance, which proves the robustness of our method.

In summary, our overall contribution is summarized as:
\begin{itemize}
	\item We propose the first content-variant reference method via knowledge distillation (CVRKD-IQA), which introduces more HQ-LQ distribution difference knowledge.
	\item With the guidance of non-aligned reference image and knowledge distillation, our model significantly outperforms existing NR/NAR-IQA methods on synthetic and authentic IQA datasets. On some occasions, our model even reaches comparable results with FR-IQA metrics.
	\item Our model can directly use content-variant HQ images for reference, which can loose the restrictions of previous pixel-aligned or content-similar reference images.
\end{itemize}

\section{Related Work}
\textbf{Image Quality Assessment.}
The target of objective IQA is to accurately acquire the consistent quality of one image with human views. According to the involvement of reference images, the objective IQA can be generally classified into three types: full-reference (FR), reduced-reference (RR), and no-reference (NR) IQA methods.

In general, FR/RR-IQA simulates the sensitivity of the human visual system to different image signals~\cite{sheikh2006image}, including information-theoretic criterion~\cite{wang2004image}, structural information~\cite{zhang2011fsim}, etc. The FR-IQA methods perform their quality measurements based on point-by-point comparisons between pixels. And FR-IQA has been widely applied as the perceptual metric for downstream tasks of image proceeding. The most commonly and widely used FR metrics are the PSNR and SSIM~\cite{wang2004image}, which are convenient for optimization. Recently, learning-based FR-IQA methods~\cite{prashnani2018pieapp, ding2021comparison} have achieved significant improvement. The most current IQT model~\cite{cheon2021perceptual} involves the visual transformer with extra quality and position embeddings to achieve the best performance for the FR-IQA task. Different from FR-IQA, the RR-IQA method~\cite{rehman2012reduced} utilizes only parts of the FR image information. Since RR-IQA has the advantages of lower calculation expense and faster speed, it's commonly applied in the image transmission system.

For NR-IQA, CNN-based methods~\cite{bosse2017deep,wu2020end,su2020blindly} have significantly outperformed handcrafted statistic-based approaches~\cite{xu2016blind} by directly extracting discriminative features from LQ images. Due to distortion diversity and content changes, the recent trend of NR-IQA~\cite{li2020norm} is to involve semantic prior information by using pretrained models on classification databases, \ie, ImageNet~\cite{deng2009imagenet}. And Su \etal~\cite{su2020blindly} propose a dynamic hyper-network to adaptively adjust the quality prediction
parameters based on image content. Recently, You \etal~\cite{you2021transformer} introduce the visual transformer for the NR-IQA task. However, FR-IQA methods tend to provide more reliable quality evaluation than NR-IQA models~\cite{zhang2011fsim}.

Since pixel-aligned FR images are not always available, DCNN~\cite{liang2016image} defines a new task named Non-aligned Reference IQA (NAR-IQA), which uses a reference image with similar scene but is not well aligned with the LQ image. Nevertheless, the images with similar scenes are still not always easy to get. Recently, Ma \etal ~\cite{ma2017dipiq} form the quality-discriminable image pairs to help rank the IQA scores, and Guo \etal~\cite{guo2021subjective} introduces the pseudo images for reference. However, those methods still need to manually form their reference images. In this paper, we attempt to use content-variant HQ images for reference.

\noindent\textbf{Knowledge Transfer via Distillation.}
Transferring knowledge from one model to another has been a long line of research. Ba and Caruana~\cite{ba2014deep} successfully increase the accuracy of a shallow neural network by training it to mimic a deeper one and penalize the difference of logits between them. Hinton et
al.~\cite{hinton2015distilling} revive this idea under the name of knowledge distillation (KD) that trains a student model to match the distribution of a teacher model. Although the KD strategy was primarily proposed for model compression~\cite{lan2018knowledge}, many recent works have extended the cross-modal distillation to multi-modal visual tasks, such as action recognition~\cite{garcia2018modality}, person re-identification~\cite{porrello2020robust} or depth estimation~\cite{gupta2016cross}, where the knowledge of different modals are transferred between different network branches. In this paper, we make the first attempt to transfer more HQ-LQ difference prior information from the FR-IQA to the NAR-IQA via KD. Experiments prove that distillation operation can further help our NAR-student achieve more accurate and stable performance.

\section{Proposed Method}
In this section, we will introduce the structure of CVRKD-IQA and explain how to transfer the distribution difference knowledge from FR-teacher to NAR-student.
\subsection{Network Architecture}
\noindent\textbf{Overall Architecture.} As shown in Fig.~\ref{network}, our model consists of two parts: FR-teacher $N_{T}$ and NAR-student $N_S$. Both of them use two types of images: the distorted LQ image $I_{LQ}$ for assessment and the HQ image $I_{HQ}$ for reference. The FR-teacher $N_{T}$ and NAR-student $N_S$ have the same structure. The only difference between them is $I_{HQ}$, where FR-teacher $N_{T}$ uses pixel-aligned $I_{FR}$ and NAR-student $N_S$ uses random non-aligned content-variant $I_{NAR}$.

Image quality is perceived by both local degradation and global information. Recently, some representative IQA methods~\cite{su2020blindly, cheon2021perceptual} usually use one local image patch as input, and average or reweight the predicted scores of each local patch to get final results. However, this operation does not make full and effective use of local-global combined information. Compared with single image patch input, multiple patches input can more effectively provide information on both local fine-grained distortion from the single patch and global coarse-grained composition cross patches at one time. Hence, our model uses 2 sets of $m$ multi-patches as inputs, \ie, $P_{LQ}=\{p_{{LQ}_i}\}(i=1,...m)$ and $P_{HQ}=\{p_{{HQ}_i}\}(i=1,...m)$, which are randomly cropped from $I_{LQ}$ and $I_{HQ}$. It should be noted that $\{P_{LQ},P_{FR}\}$ are still pixel-aligned for the FR-teacher.

To combined the advantages of NR-IQA and FR-IQA methods, our model attempts to mine the local-global combined features from the LQ image itself and HQ-LQ distribution difference. Moreover, the multi-scale feature extraction should also be conducted to better describe the local distortion. To achieve those, we design three modules for our model: (1) the multi-scale local feature extractor; (2) dual cross-patch mixing encoders; (3) a full-connected regressor;

\noindent\textbf{Multi-scale Local Feature Extractor.} First, following~\cite{li2020norm,cheon2021perceptual}, the pretrained CNN backbone on the image classification task is applied as the perceptual feature extractor. Thus, we use the pretrained ResNet50~\cite{he2016deep} on ImageNet~\cite{deng2009imagenet} to process input patches $\{P_{LQ},P_{HQ}\}$. Since features from different scale layers are important to capture local distortions~\cite{su2020blindly,guo2021iqma}, we design a multi-scale feature extractor. Specifically, four scale features from conv2\_9, conv3\_12, conv4\_18, conv5\_9 layers of ResNet50 are processed by $1\times 1$ convolution and global average pooling. Those four feature maps with the same size $([m, 64, 7, 7])$ are concatenated as $f_m$ in channel-wise $([m, 256, 7, 7])$ to describe local distortions.

\noindent\textbf{Dual Cross-patch Mixing Encoder.} Then, we use cross-patch mixing encoders to extract the self-perceived feature $f_{LQ}$ and the HQ-LQ difference-aware features $f_{HQ-LQ}$, respectively. To effectively explore local-global combined information from multi-patches input, we build our encoders with the classic MLP-mixer~\cite{tolstikhin2021mlp}, which has a simpler architecture and faster speed than the visual transformer~\cite{vaswani2017attention}. Different from the original MLP-mixer~\cite{tolstikhin2021mlp} for classification tasks fed with spatial image tokens, our encoders operate on the multi-scale features $f_m$ extracted from multi-patches input. Each MLP module consists of two blocks: the first one is patch-mixing MLP block, which exchanges inner information between transposed local features of multi-patches; the next one is channel-mixing MLP block, which allows global information communication between multi-patches and multi-scales. Since mining the distribution difference between $I_{LQ}$ and
$I_{HQ}$ is much more difficult than $I_{LQ}$ perceive feature extraction, we design deeper encoder $E_{HQ-LQ}$ with 18 stacked MLP modules and the encoder $E_{LQ}$ uses only 9 stacked MLP modules. The final layer normalization and global average pooling convert feature maps to vectors. Two cross-patch vectors ([256, 1]) from the dual-path encoder are concatenated as ([512, 1]) for quality regression prediction.

\noindent\textbf{Regressor for Quality Prediction.}
Since the regressor is simply mapping the output vectors of the dual-path encoder to labeled quality scores, we design a small network for faster quality prediction. The regressor consists of two fully-connected layers with 512-256, 256-1 channels to predict the final quality score of the input LQ image.

\subsection{Knowledge Distillation from FR-IQA to NAR-IQA}
Considering that our goal is to transfer more HQ-LQ distribution knowledge, and better constrain NAR-student for useful HQ-LQ distribution difference representation, we perform the distillation operation between the difference-aware encoders $E_{HQ-LQ}$ of FR-teacher and NAR-student.

To obtain a well-performed FR-teacher, we do not jointly train FR-teacher and NAR-student, but apply an offline distillation scheme. First, we randomly crop two multi-patch sets
$\{P_{LQ},P_{FR}\}$ from the LQ-FR image pair $\{I_{LQ},I_{FR}\}$ as the input. The FR-teacher $N_{T}(\cdot;\theta_1)$ is optimized by $L_1$ loss between predicted score $\hat y_t$ and ground-truth $y$ as:
\begin{align}\label{teacher label loss}
	L_{T_l} = \frac{1}{N}\sum_{i=1}^N||y_i-N_T(P_{LQ}^{(i)},P_{FR}^{(i)};\theta_1)||_1.
\end{align}

Then, we fix the parameters of the trained FR-teacher. The NAR-student is supervised by the guide of FR-teacher and human labeled scores in the second step of training. Except the paired $\{P_{LQ},P_{FR}\}$ for FR-teacher input, NAR-student should also be fed with the non-aligned $\{P_{LQ},P_{NAR}\}$, where $P_{NAR}$ consists of $m$ randomly cropped patches from another non-aligned reference HQ image. We attempt to transfer more prior knowledge of HQ-LQ distribution difference from FR-teacher to NAR-student. Hence, all 18 inner features $F_T=\{f_{T_j}\}(j=1,2,...18)$ of the difference-aware encoder of FR-teacher are applied to guide the training of NAR-student. The $L_2$ loss is used as the distillation loss $L_{S_d}$ to transfer knowledge to corresponding layer features $F_S=\{f_{S_j}\}(j=1,2,...18)$ of NAR-student:
\begin{align}\label{distillation loss}
	L_{S_d} = \frac{1}{N}\sum_{i=1}^N\sum_{j=1}^{K=18}||f_{T_j}^{(i)}-f_{S_j}^{(i)}||_2.
\end{align}
Except distillation loss, the label loss $L_{S_l}$ between predicted results $\hat y_s$ and labeled ground-truth $y$ is also applied to optimize the NAR-student $N_{S}(\cdot;\theta_2)$:
\begin{align}\label{student label loss}
	L_{S_l} = \frac{1}{N}\sum_{i=1}^N||y_i-N_S(P_{LQ}^{(i)},P_{NAR}^{(i)};\theta_2)||_1.
\end{align}
And the final loss $L_{S}$ for NAR-student is combined by the distillation loss $L_{S_d}$ in Eq.~\ref{distillation loss} and label loss Eq.~\ref{student label loss} as:
\begin{align}
	L_{S} = L_{S_d} + L_{S_l}.
\end{align}
With the guidance of knowledge distillation, our NAR-student effectively learns more HQ-LQ difference knowledge and keeps the stability with different NAR images. In real scenarios, when the pixel-aligned FR image is unavailable but HQ images are easy to get, our NAR-student can directly use any non-aligned HQ image for reference.

\begin{table*}[tbp]\footnotesize
	\renewcommand\tabcolsep{1pt} 
\centering
\begin{tabular}{l|l|ccc|ccc|ccc|ccc}
	\toprule
    \multirow{2}*{IQA Type}&\multirow{2}*{Method}&\multicolumn{3}{c|}{LIVE}&\multicolumn{3}{c|}{CSIQ}&\multicolumn{3}{c|}{TID2013}&\multicolumn{3}{c}{KonIQ-10K}\\
		&&SRCC&PLCC&KRCC&SRCC&PLCC&KRCC&SRCC&PLCC&KRCC&SRCC&PLCC&KRCC\\
		\hline
		\hline
		\multirow{8}{*}{FR-IQA}&PSNR&0.873&0.865&0.680&0.810&0.819&0.601&0.687&0.677&0.496&-&-&-\\
		&MAD~\cite{larson2010most}&0.967&0.968&0.842&0.947&0.950&0.797&0.781&0.827&0.604&-&-&-\\
		&WaDIQaM-FR~\cite{bosse2017deep}&0.947& 0.940&0.791&0.909& 0.901&0.732&0.831& 0.834&0.631&-&-&-\\
		&PieAPP~\cite{prashnani2018pieapp}&0.919&0.908&0.750&0.892&0.877&0.715&0.876&0.859&0.683&-&-&-\\
		&LPIPS~\cite{zhang2018unreasonable}&0.932&0.934&0.765&0.876&0.896&0.689&0.670&0.749&0.497&-&-&-\\
		&DISTS~\cite{ding2021comparison}&0.954&0.954&0.811&0.929&0.928&0.767&0.830&0.855&0.639&-&-&-\\
		&IQT~\cite{cheon2021perceptual}&0.970&-&0.849&0.943&-&0.799&\textbf{0.899}&-&\textbf{0.717}&-&-&-\\
		&Our FR-teacher&\textbf{0.973}&\textbf{0.969}&\textbf{0.853}&\textbf{0.964}&\textbf{0.964}&\textbf{0.829}&0.890&\textbf{0.886}&0.698&-&-&-\\
		\hline
		\hline
		\multirow{5}{*}{NR-IQA}&CNNIQA~\cite{kang2014convolutional}&0.653&0.656&0.485&0.649&0.660&0.482&0.476&0.404&0.283&0.278&0.285&0.183\\
		&WaDIQaM-NR~\cite{bosse2017deep}&0.855&0.855&0.656&0.716&0.750&0.527&0.585&0.610&0.416&0.382&0.386&0.261\\
		&HyperIQA~\cite{su2020blindly}&0.908&0.903&0.730&0.802&0.858&0.611&0.686&0.721&0.490&0.332&0.338&0.233\\
		&TRIQ~\cite{you2021transformer}&0.909&0.910&0.729&0.807&0.862&0.615&0.684&0.731&0.500&0.371&0.371&0.259\\
		&LinearityIQA~\cite{li2020norm}&0.910&0.906&0.738&0.815&0.873&0.629&0.688&0.694&0.491&0.361&0.361&0.254\\
		\hline
		\multirow{4}{*}{NAR-IQA}&DCNN~\cite{liang2016image}&0.752&0.756&0.594&0.721&0.716&0.583&0.473&0.492&0.346&0.258&0.256&0.147\\
		&WaDIQaM~\cite{bosse2017deep}-NAR w/ KD& 0.897& 0.894& 0.707& 0.799& 0.851& 0.613& 0.670 & 0.694& 0.493& 0.362& 0.364& 0.258\\
		&IQT~\cite{cheon2021perceptual}-NAR w/ KD& 0.908& 0.906& 0.728& 0.802& 0.860& 0.624& 0.680& 0.707& 0.499&0.372 & 0.372& 0.269\\
		&Our NAR-student&\textbf{0.913}&\textbf{0.917}&\textbf{0.748}&\textbf{0.829}&\textbf{0.872}&\textbf{0.655}&\textbf{0.691}&\textbf{0.733}&\textbf{0.501}&\textbf{0.416}&\textbf{0.413}&\textbf{0.287}\\
		\bottomrule

\end{tabular}
\caption{Model comparisons on synthetic LIVE, CSIQ, TID2013, and authentic KonIQ-10K when training on synthetic Kaddid-10K. We also extend two FR-IQA methods (WaDIQaM, IQT) to NAR-IQA via knowledge distillation (KD). It's clear that our NAR-student can outperform all NR/NAR-IQA methods, especially on the large-scale authentic KonIQ-10K with real unknown distortions. On TID2013, our NAR-student reaches comparable and even better performance than PSNR and LPIPS.}
\label{SOTA}
\end{table*}

\section{Experiments}
\subsection{Experimental setting}
In this paper, all comparisons of FR/NR/NAR-IQA methods and ablation studies follow this setting.

\noindent\textbf{Datasets.} For IQA training datasets, we follow~\cite{cheon2021perceptual} to choose the commonly used synthetic Kaddid-10K~\cite{lin2019kadid}, which contains 10125 LQ-FR pairs. The cross-dataset evaluations are conducted on 3 synthetic datasets, \ie, LIVE~\cite{sheikh2006statistical}, CSIQ~\cite{larson2010most}, TID2013~\cite{ponomarenko2015image}, which separately contains 779, 886 and 3000 LQ-FR pairs with traditional distortions. Moreover, we also evaluate on large-scale authentic KonIQ-10K dataset~\cite{hosu2020koniq}, containing 10073 real-distorted LQ images without FR images. Except for IQA datasets, our NAR-student still need non-aligned HQ reference images. The 900 training and 100 testing HQ images of DIV2K\_HR dataset~\cite{agustsson2017ntire} are randomly sampled at the training and testing stages of NAR-student.

\noindent\textbf{Evaluation Criterias.} The Spearman’s rank order correlation coefficient (SRCC), Pearson’s linear correlation coefficient (PLCC) and Kendall rank order correlation coefficient (KRCC) are employed to measure prediction monotonicity and prediction accuracy. The higher value indicates better performance. For PLCC, the logistic regression correction is also applied according to~\cite{antkowiak2000final}.

\noindent\textbf{Implementation Details.} Data augmentation including horizontal flip and random rotation is applied during the training. All patches are randomly cropped from the RGB image. The batch size $b$ is set as 32. The input patch number $m$ is set as 10 and the patch size is set as $224\times 224 \times 3$ to cover more local-global combined information. The number $k$ of distilled layers in the encoder $E_{HQ-LQ}$ is set to 18. Moreover, the initial learning rate $\alpha$ is $2\times 10^{-5}$ and the ADAM optimizer with weight decay $5\times 10^{-4}$ is applied. All the experiments were conducted on NVIDIA Tesla-V100 GPUs.

\subsection{Comparisons with the State-of-the-art Methods}
Here, we will present the accuracy and generalization comparisons between our model and existing FR/NR/NAR-IQA methods. Specifically, our FR-teacher and NAR-student are separately compared with FR-IQA SOTAs \ie,~\cite{ding2021comparison, cheon2021perceptual} and recent best performed NAR/NR-IQA SOTAs \ie,~\cite{liang2016image, li2020norm}.
Moreover, we also extend two FR-IQA methods (WaDIQaM~\cite{bosse2017deep}, IQT~\cite{cheon2021perceptual}) to NAR-IQA via knowledge distillation. Specifically, we use the pretrained WaDIQaM and IQT as teachers under FR-IQA settings. And we obtain the corresponding student models by changing the reference input from FR images to NAR images. Following our strategy, the knowledge distillation is also applied in those models and we get the WaDIQaM-NAR and IQT-NAR w/ KD. Since we follow the commonly used experimental settings of FR-IQA methods~\cite{ding2021comparison,cheon2021perceptual}, we directly use those published FR-IQA results. For fair comparisons between FR/NR/NAR methods, we retrain those NR/NAR-IQA SOTAs with the same experimental setting as ours.

The results of the four datasets are shown in Table~\ref{SOTA}. For the FR-IQA setting, since the authentic Kaddid-10K doesn't provide FR images, the FR-IQA comparisons are only conducted in 3 synthetic datasets. It can be seen that our FR-teacher outperforms all FR-IQA methods on LIVE and CSIQ, and it is also ranked in the top two in all benchmarks with the marginal gap on larger TID2013. Hence, the trained FR-teacher is good enough as the distillation teacher. When pixel-aligned FR images are not provided, our NAR-student outperforms existing NR/NAR-IQA models on all 4 testsets, which shows that the proposed strategy can improve the IQA performance.
What's more, the comparisons about distilled WaDIQaM-NAR and IQT-NAR further prove this point. It should be noted that although trained on the synthetic Kaddid-10K, our NAR-student achieves significant improvement than NR/NAR-IQA SOTAs on the authentic KonIQ-10K. Moreover, on the synthetic TID2013, our NAR-student reaches comparable and even better performance than the commonly used FR-IQA methods, such as PSNR and LPIPS~\cite{zhang2018unreasonable}.

\subsection{Runtime vs. Performance}
\begin{figure}[htbp]
\centering
\includegraphics[width=6.0cm]{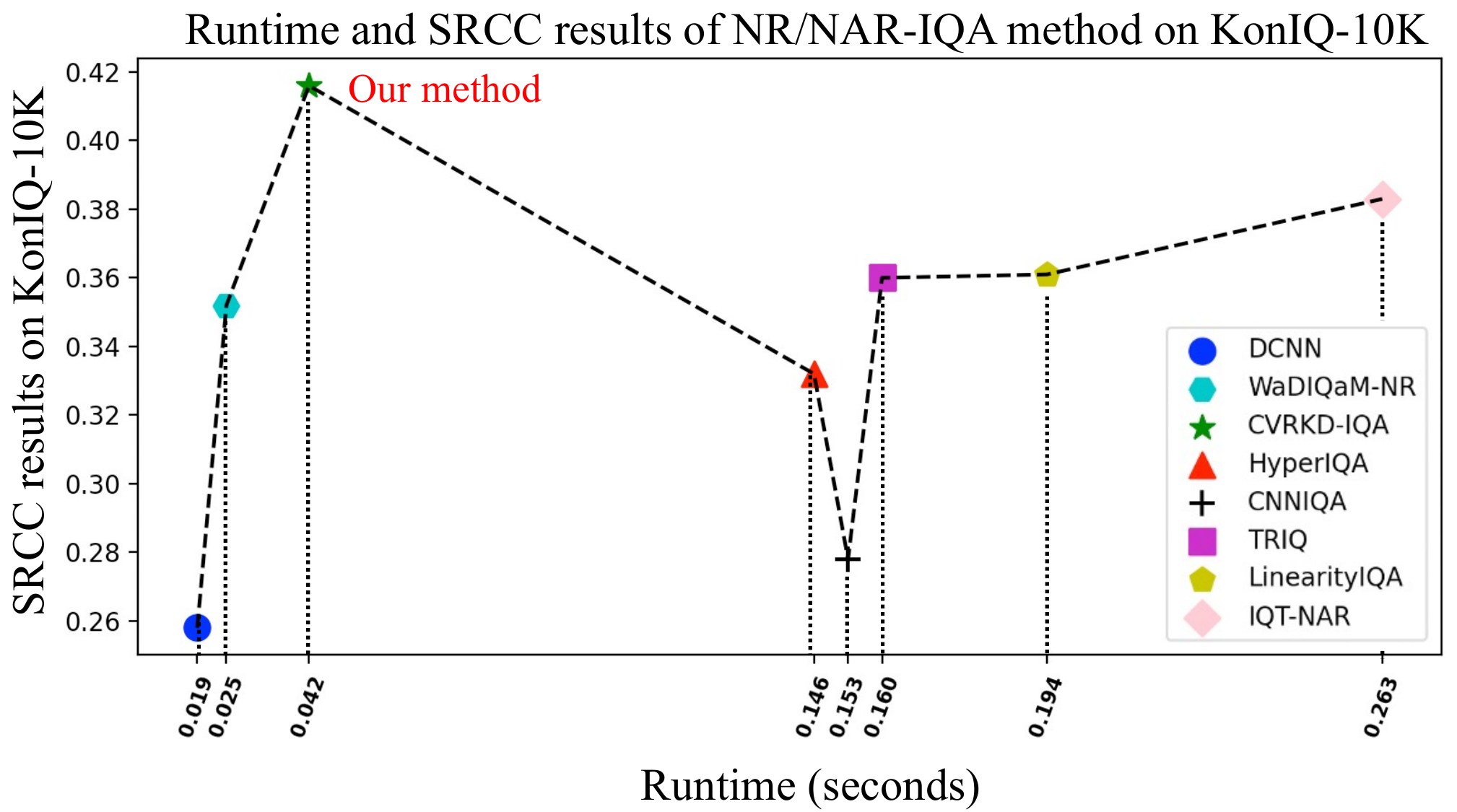}
\caption{Runtime vs. performance of NR/NAR-IQAs on the real-distorted KonIQ-10k dataset with the Tesla-V100 GPU.}
\label{runtime}
\end{figure}

To compare the efficiency of our NAR-IQA model with other NR/NAR methods in the inference stage, we report the average runtime of IQA for a distorted image with the number of patches $m=10$ in Fig.~\ref{runtime}. On real-distorted KonIQ-10K, our CVRKD-IQA significantly outperforms NR/NAR-IQA SOTAs and satisfies the real-time requirement (about 24 images per second), while transformer-based TRIQ~\cite{you2021transformer} and IQT-NAR~\cite{cheon2021perceptual} cost much more inference time. All experiments were conducted on NVIDIA Tesla-V100 GPU.

\subsection{Ablation Study}
\noindent\textbf{Effect of Knowledge Distillation (KD) and Non-aligned Reference Images (NAR).}
First, we separately analyze the effects of knowledge distillation (KD) and non-aligned reference (NAR) images in Table~\ref{configurations}. It should be noted that NAR is the pre-condition of KD. If the NAR image is not provided, the NAR-student cannot mine the HQ-LQ distribution difference, thus cannot learn the transferred knowledge from the FR-teacher. Hence, we evaluate 3 types of KD and NAR configurations. Except SRCC metrics, we also present the standard deviations (Std) of 10 SRCC results tested with different misaligned reference images. From results in Table~\ref{configurations}, we can make the following analyses:
\begin{itemize}
	\item[-] We first remove the difference-aware encoder to train NR-student baseline under NR-IQA setting without NAR or KD. It's clear the NR-student baseline achieves worse performance, especially in real-distorted KonIQ-10K.
	\item[-] When NAR images are available, the NAR-student w/o KD benefits from the HQ-LQ distribution difference to outperform the NR-student baseline. However, the performance of NAR-student w/o KD is the most unstable with the highest Std values. It means various contents of different NAR images increase the training difficulty.
	\item[-] When provided with more HQ-LQ difference knowledge from the FR-teacher by KD, our final NAR-student achieves the best performance, especially the 33\% SRCC improvements than NR-student baseline in KonIQ-10K. Moreover, Std results of our final NAR-student decreased to 0.004, which proves the great importance of KD to stabilize the NAR-IQA performance.
\end{itemize}

\begin{table}[tbp]\footnotesize
	\renewcommand\tabcolsep{1pt} 
	\centering
	\begin{tabular}{l|cc|c|c}
		\toprule
		\multirow{2}{*}{Model}&\multicolumn{2}{c|}{Configs}&TID2013&KonIQ-10K\\
		& NAR&KD&SRCC $\uparrow$ $\pm$ Std $\downarrow$&SRCC $\uparrow$ $\pm$ Std $\downarrow$\\
		\midrule
		\midrule
		NR-student baseline & \texttimes& \texttimes& 0.631 $\pm$  \textbf{0.002}& 0.317 $\pm$ \textbf{0.003}\\
		\midrule
		NAR-student w/o KD & \checkmark& \texttimes& 0.679 $\pm$ 0.056& 0.352 $\pm$ 0.072\\
		NAR-student w/ KD & \checkmark& \checkmark& \textbf{0.691} $\pm$ 0.003& \textbf{0.416} $\pm$ 0.004\\
		\bottomrule
	\end{tabular}
	\caption{SRCCs and the standard deviations (Std) of our student with different configurations of knowledge distillation (KD) and non-aligned reference (NAR) images.}
	\label{configurations}
\end{table}

\begin{figure}[tbp]
\centering
\includegraphics[width=8.3cm]{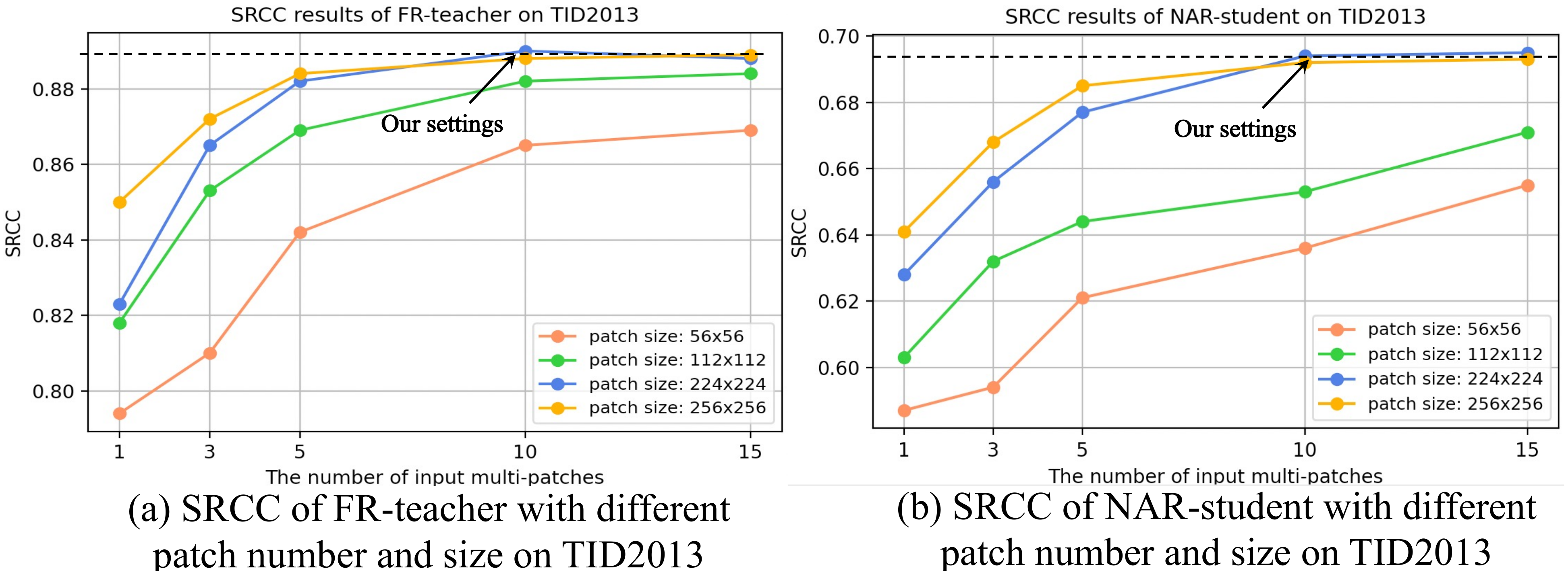}
\caption{SRCC results with different multi-patch numbers and sizes on TID2013. It's clear that larger patch number and size can capture more local-global information.}
\label{patch}
\end{figure}

\noindent\textbf{Effectiveness of Multi-patches.}
To make full and effective use of local-global combined information, our method directly processes multi-patches and fuses the cross-patch features via the MLP-mixer.
As shown in Fig.~\ref{patch}, we gradually increase the patch number (1, 3, 5, 10, 15) and the patch size (56, 112, 224, 256) to analyze the effects of multi-patches. It's clear that both FR-teacher and NAR-student benefit from larger patch number and size, because they can capture more local-global information to better describe the full-image quality. Considering the trade-off between inference efficiency and performance, the patch number is set to $10$ and the patch size is set to $224\times 224$.

\begin{figure}[tbp]
\centering
\includegraphics[width=8cm]{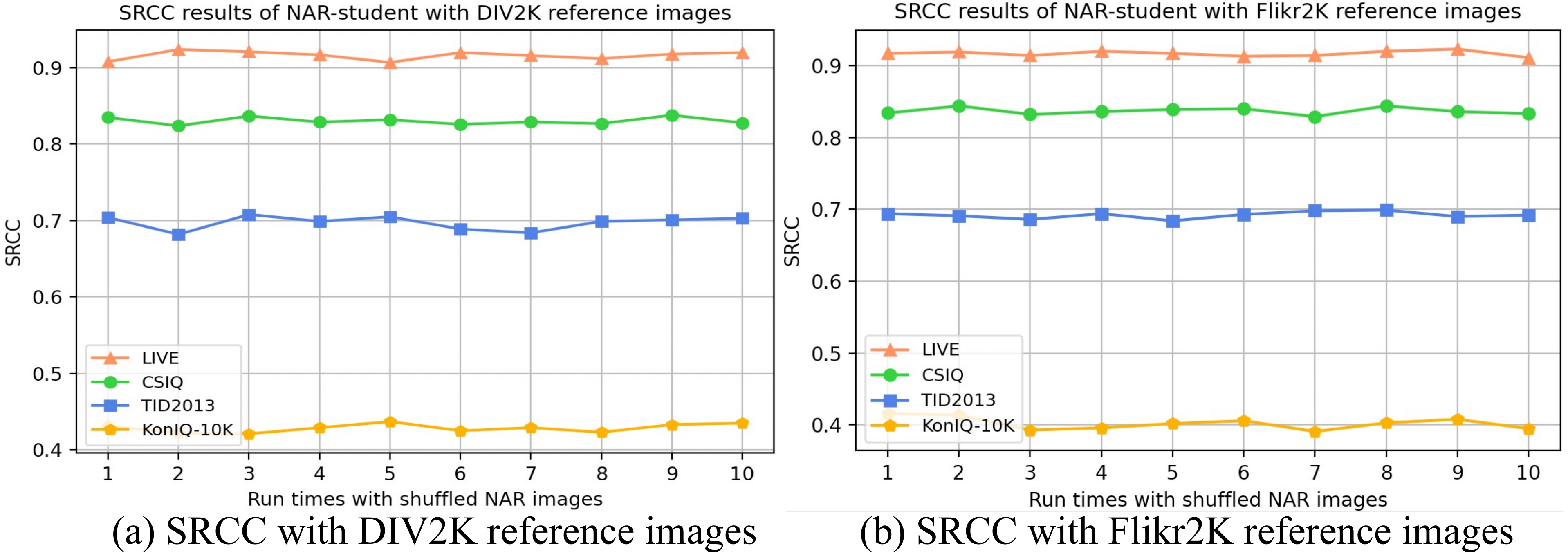} 
\caption{Stability evaluation of our NAR-student. (a)(b) show SRCC results of 10 times tests using randomly shuffled NAR images from DIV2K and Flikr2K, respectively.}
\label{stable}
\end{figure}

\noindent\textbf{Stability about Non-aligned Reference HQ Images.}
To further demonstrate the stability of our model when using content-variant HQ images for reference, we  evaluate our NAR-student with more various HQ images. Specifically, not only DIV2K, we involve 2650 HQ images of Flikr2K~\cite{timofte2017ntire} as another non-aligned reference dataset. Since we randomly sample the HQ reference image for each LQ assessment, HQ images of each round are shuffled. Therefore, we also present results across 10 times. As shown in Fig.~\ref{stable}, we can make the following analyses:
\begin{itemize}
	\item[-] As shown in Fig.~\ref{stable}(a)(b), our NAR-student achieves relatively stable performance when using shuffled NAR images across 10 times on both DIV2K and Flikr2K.
	\item[-] As shown in the comparisons between Fig.~\ref{stable}(a) and (b), our NAR-student also produces relatively similar SRCC results between DIV2K and Flikr2K.
	\item[-] Those observations demonstrate that our NAR-student is stable and robust to content-variant NAR images.
\end{itemize}

%

\begin{figure}[htbp]
\centering
\includegraphics[width=7.8cm]{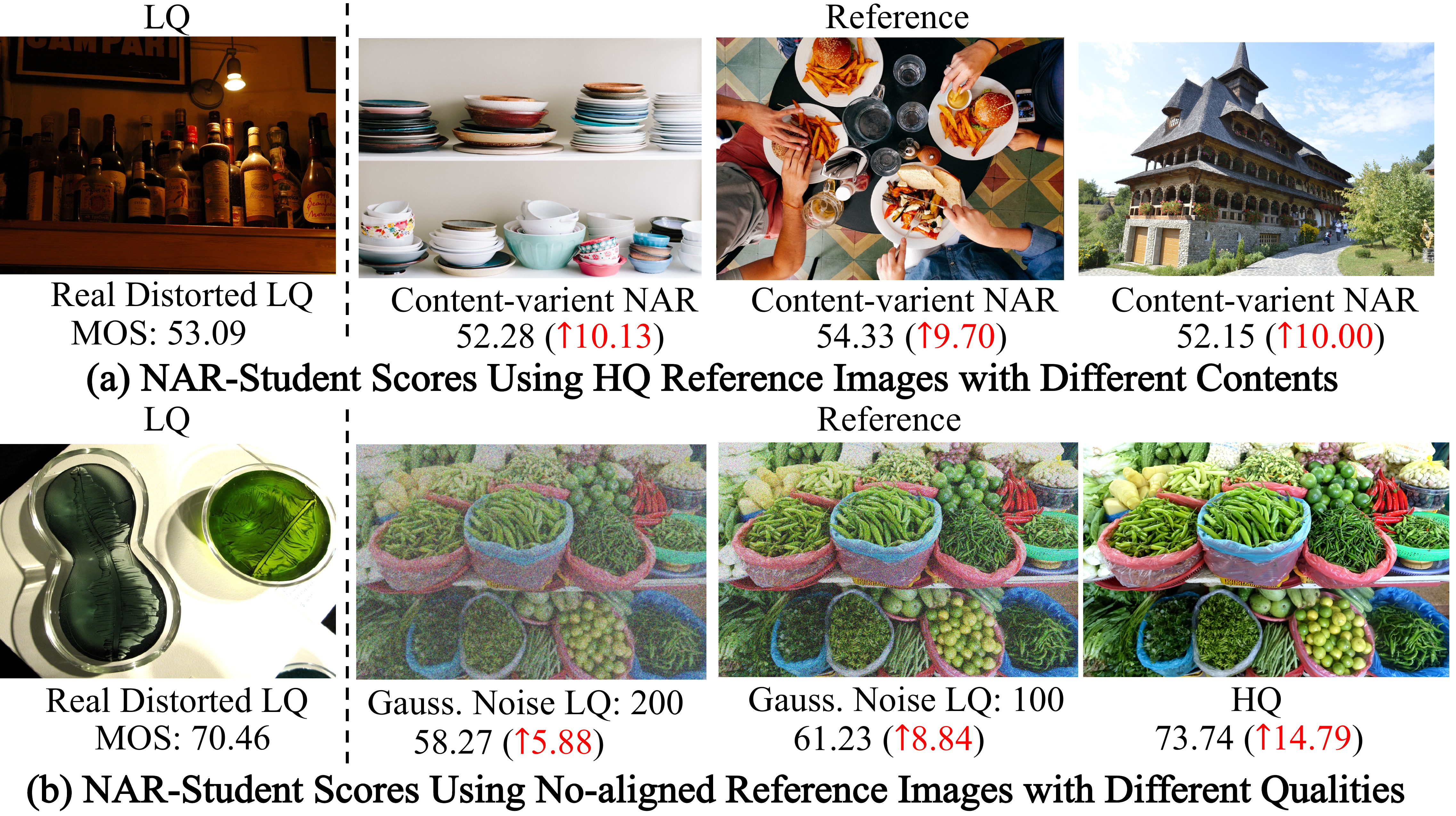} 
\caption{\textbf{Real-data examples} on KonIQ-10K, where the NAR-student uses reference images with (a) different contents and (b) different qualities.
All scores are rescaled in [0, 100]. The red numbers are the decrease of the MOS error compared to the no-reference baseline.}
\label{visual}
\end{figure}

\begin{table}[t]\footnotesize
	\renewcommand\tabcolsep{2pt} 
	\centering
	\begin{tabular}{l|c|c|c}
		\toprule
		\multirow{2}{*}{Model + Input Reference Image}&LIVE&CSIQ&TID2013\\
		&SRCC&SRCC&SRCC\\
		\midrule
		\midrule
		FR-teacher + Pixel-aligned FR & 0.973& 0.964& 0.890\\
		\midrule
		NAR-student + Pixel-aligned FR& 0.958& 0.937& 0.846\\
		NAR-student + Content-similar NAR& 0.931& 0.862& 0.720\\
		NAR-student + Content-variant NAR& 0.913& 0.829& 0.691\\
		\bottomrule
	\end{tabular}
	\caption{SRCC results using HQ reference images with different contents. For clear comparisons, we add the FR-teacher with pixel-aligned FR image as the upper-bound. It's clear that more aligned HQ images produce better results.}
	\label{contents}
\end{table}

\noindent\textbf{Evaluation on Reference Image with Different Content.}
Now, there are 3 types of reference images: the pixel-aligned FR image, the NAR image with similar content and the random content-variant NAR image. How can we choose them properly based on content? Hence, we evaluate the distilled NAR-student with different types of HQ reference contents. Note that we follow ~\cite{liang2016image} to synthesize the content-similar NAR image by applying affine transform to FR images (random scaling factor $s$ and rotation $\theta$ from $[0.95,1.05]$ and $[-5^{\circ},5^{\circ}]$). From results in Table~\ref{contents} and Fig.~\ref{visual}(a), we can make the following analyses:
\begin{itemize}
		\item[-] As shown in the first two lines of Table~\ref{contents}, although the NAR-student is trained with NAR settings, it reaches comparable results with
	   FR-teacher when using pixel-aligned FR images. This proves the NAR-student has learned transferred knowledge from the FR-teacher.
		\item[-] As shown in the last two lines of Table~\ref{contents}, the performance of content-variant NAR images is slightly lower than content-similar NAR images. Since the stable and promising performance of our method has been proved in Fig.~\ref{visual}(a) and Fig.~\ref{stable}, we can use random HQ images for reference when content-similar images are unavailable.
		\item[-] In real scenarios, we should choose the HQ reference image with aligned content as much as possible.
	\end{itemize}

\noindent\textbf{Evaluation on Reference Image with Different Quality.}
Although the HQ images are easy to obtain, we should still evaluate our NAR-student on NAR images with different qualities. Specifically, we first use various distorted reference images of synthetic TID2013 and authentic KonIQ-10K. Moreover, we choose 3 typical distortions to generate distorted reference images from DIV2K\_HQ, \eg, $\times$2 downsample, random Gaussian noise with levels: [0,10], random JPEG compression with qualities: [0,10]. From results in Table~\ref{quality} and  Fig.~\ref{visual}(b), we can make the following analyses:
\begin{itemize}
		\item[-] As shown in Table~\ref{quality} and Fig.~\ref{visual}(b), using reference images with higher quality can produce better results.
		\item[-] As shown in the first two examples of Fig.~\ref{visual}(b), using severely distorted LQ images for reference just brings marginal improvements than the no-reference baseline.
		\item[-] In real scenarios, we should choose the NAR image with high-quality as much as possible. Since the HQ images are easy to get, our NAR-student can directly use random obtainable HQ images for reference.
\end{itemize}
\begin{table}[tbp]\footnotesize
	\renewcommand\tabcolsep{1pt} 
	\centering
	\begin{tabular}{l|l|cc|cc}
		\toprule
		\multirow{2}{*}{Reference}&\multirow{2}{*}{Distortion Type} &\multicolumn{2}{c|}{TID2013}&\multicolumn{2}{c}{KonIQ-10K}\\
		&&SRCC&PLCC&SRCC&PLCC\\
		\midrule
		\midrule
	 	KonIQ-10K & Authentic Distortions& 0.671& 0.711 & 0.392&0.393\\
	  TID2013 & Synthetic Distortions& 0.683& 0.722& 0.394&0.395\\
		\midrule
		\multirow{3}{*}{DIV2K} & Gauss. Noise: [0, 10] & 0.665& 0.704& 0.392&0.390\\
		& JPEG Level: [0, 10] & 0.668& 0.702& 0.401&0.401\\
		& Downsample: $\times$2& 0.687& 0.721& 0.408& 0.407\\
		\midrule
		\multicolumn{2}{c|}{DIV2K\_HQ}& \textbf{0.691}&\textbf{0.733}&\textbf{0.416}&\textbf{0.413}\\
		\bottomrule
	\end{tabular}
	\caption{The SRCC and PLCC results of our NAR-student with different qualities NAR images. The NAR-student are fixed, and we only change the types of reference images. It's clear NAR images with higher quality produce better results. }
	\label{quality}
\end{table}

\section{Conclusion}
In this paper, we investigate the image quality assessment (IQA) problem with non-aligned reference (NAR) images. We propose the first content-variant NAR-IQA method via knowledge distillation, namely CVRKD-IQA. Our model uses various NAR images to introduce prior distributions of HQ images. The knowledge distillation further transfers more HQ-LQ distribution difference knowledge from the FR-teacher to the NAR-student and stabilizes IQA performance. We also use the multiple patches input to fully and effectively mine the multi-scale and local-global combined features. Extensive experiments have demonstrated that our CVRKD-IQA significantly outperforms existing NR/NAR-IQA methods, even reaches comparable performance with commonly used FR-IQA metrics. Evaluations with different NAR images also prove the relative robustness of our model, which can support more IQA applications with randomly obtainable HQ images. Moreover, the reference images with higher-quality and more aligned content produce better results. In future work, we will further explore more novel NAR-IQA architecture and knowledge distillation strategy.

{\small
		\bibliographystyle{aaai21}
		\bibliography{ref}
	}
\end{document}